\documentclass[letterpaper]{article}
\usepackage[preprint]{aaai2027}
\usepackage[hyphens]{url}
\usepackage{graphicx}
\usepackage{natbib}
\usepackage{caption}
\usepackage{algorithm}
\usepackage{algorithmic}
\usepackage{booktabs}
\usepackage{array}
\usepackage{amsmath}
\usepackage{amssymb}
\usepackage{amsthm}
\theoremstyle{definition}
\newtheorem{definition}{Definition}

\usepackage{xcolor}

\definecolor{linkblue}{RGB}{0,0,190}

\title{Measuring Semantic Abstractness of SAE Features via Nonlocality}

\author{
    Chuqiao Lin\textsuperscript{\rm 1},
    Shivaji Sondhi\textsuperscript{\rm 1},
    Xiao-Liang Qi\textsuperscript{\rm 2}
}
\affiliations{
    \textsuperscript{\rm 1}Rudolf Peierls Centre for Theoretical Physics, University of Oxford\\
    \textsuperscript{\rm 2}Leinweber Institute for Theoretical Physics, Stanford University\\
    chuqiao.lin@physics.ox.ac.uk
}

\begin{document}
\maketitle

\begin{abstract}
    Sparse autoencoders (SAEs) have helped uncover mechanistic explanations for LLM behaviours such as reasoning, jailbreaking etc., via understanding the corresponding task-relevant and causally effective features. To evaluate such mechanistic explanations, downstream studies must distinguish surface lexical features from genuinely high-level ones. However, neither an autointerp-based semantic description nor causal steering utility fully resolves the abstraction level of a feature. To this end, we introduce \emph{Feature Nonlocality} (FNL), defined as the entropy of the normalized per-position influence on an SAE feature's activation. We report that FNL correlates with existing LLM-based proxy metrics of feature semantic abstractness, and successfully distinguishes context-dependent reasoning features from token-driven ones, correctly assigning the higher FNL to the contextual feature in $73$--$84\%$ of randomly drawn pairs that consist of one contextual and one token-level feature.
    We demonstrate two downstream applications. We audit SAE-based features used for jailbreak mitigation and find surprisingly that most effective features are positional features with low FNL rather than genuinely recognizing harmful intents.
    We report that steering high-FNL features in DeepSeek-R1-Distill-Llama-8B improves MATH-500 accuracy by $4.6$ points over the unsteered model and outperforms steering low-FNL features, though the gains are model-specific. We conclude that FNL provides an LLM-independent, label-free, correlational witness of the abstraction level of an SAE feature, with applications in evaluating mechanistic explanations as well as selecting features for downstream interventions.
\end{abstract}

\section{Introduction}

As modern Large Language Models become increasingly capable and demonstrate research-level intelligence \cite{alon2026remarksdisproofunitdistance}, it remains a central challenge to understand the mechanisms of their emergent capabilities, thereby controlling their behavioural pattern to mitigate safety risks. To this end, Mechanistic Interpretability (MI) applies probes and interventions to internal activations of Large Language Models (LLMs) at inference time, aiming at microscopic interpretation and control of their behaviour \cite{sharkey2025openproblemsmechanisticinterpretability}. Much work in MI operates under the Linear Representation Hypothesis \cite{park2024linear,elhage2022superposition}: the goal is to extract human-interpretable, causally relevant \textit{features}, i.e., linear directions in the hidden activation spaces that allow behavioural editing upon interventions at test time. Such pipelines have successfully identified interventions that reliably modify truthfulness \cite{li2024truthful}, refusal \cite{arditi2024refusal}, reasoning \cite{galichin2025reasoning}, and persona \cite{chen2025persona}, etc., without updating model weights.

To discover and extract features, Sparse Autoencoders (SAEs) offer an appealing, unsupervised machinery. By reconstructing hidden activations with an overcomplete, sparse dictionary, SAEs aim to separate superposed variables into individually monosemantic, human-interpretable features \cite{cunningham2024sparse,gao2024scalingevaluatingsparseautoencoders}. When studying target behaviour such as reasoning/jailbreaking with SAEs, previous works filter useful features from the full dictionary by (1) gating their activations on a specific subset of relevant token-cues \cite{galichin2025reasoning,fang2026controllablellmreasoningsparse}; (2) curating contrastive datasets to find features activating on the positive examples but not the negative ones \cite{assogba2026sae}; or (3) relying on LLM-generated natural language interpretations based on the text samples with significant activation of a given feature\cite{bills2023language}. Such feature selection methods are then complemented with steering experiments to confirm their behaviour editing power upon causal interventions \cite{arad2025outputscore}.

Importantly, successful \textit{control} of a target behaviour by intervening on a feature does not imply understanding of the underlying \textit{mechanism} \cite{makelov2023subspacelookingforinterpretability}. For example, if steering on a feature reliably induces backtracking, the feature can either genuinely represent detection of uncertainty, or simply boost the probability of emitting the token ``Wait'' \cite{muennighoff2025wait,ward2025reasoningfinetuningrepurposeslatentrepresentations}. It is therefore important to distinguish token-level, ``wrapper'' features from genuine high-level features in order to support a microscopic mechanistic understanding. Current feature selection methods, however, often do not distinguish mechanistic complexity of features. In particular, token-level cues and contrastive-dataset-based filtering do not rule out surface-level lexical/stylistic features. Auto-interp-based filtering often suffers from the brittleness of LLM-generated interpretations \cite{huang2023critiqueautointerp}.

To supply a quantitative witness of features' mechanistic complexity, we propose \textit{feature nonlocality} (FNL). Concretely, FNL is measured using a backward pass to see which positions across the input sequence are causally related to the feature's activation, which intuitively characterizes the \textit{contextual reach} of a feature. Token-level features (e.g.\ bigram statistics) depend on local prefix cues and have low nonlocality, while abstract, topical features draw on contextual evidence spread broadly across the window and have high nonlocality. We note that the definition of FNL takes inspiration from holographic duality \cite{maldacena1999large}, which describes a duality between a quantum gravity theory and a quantum field theory at lower dimension. Degrees of freedom at different locations in the quantum gravity theory (usually called the "bulk theory") correspond to those in the quantum field theory at different length scales. The SAE features are the analog of bulk theory degrees of freedom, and FNL measures its "location" which corresponds to nonlocality in time direction. Here we view the transformer model as the "time evolution" in the token vector space \cite{rigolette2025mathtransformers}.

We begin by confirming that FNL measures features' intrinsic properties. Across the input distributions we study (WikiText, GSM8K, Code-Python), per-feature FNL rankings remain stable: we report this for Gemma-2-2B in the main text, and for Llama-3-8B and Qwen3-8B in the supplementary material. We also report that FNL respects the feature geometry, in the sense that features with similar decoding vectors (i.e. with similar mechanistic functions on the write-side) share similar ranges of nonlocality. Next, we test whether FNL tracks the abstraction level of features using complementary evidence. First, we match FNL statistics with their auto-interp descriptions and find consistency between conceptual abstraction and higher FNL. Second, in the context of reasoning, we anchor on two independent proxy measures developed in \cite{ma2026injection}: for a given feature, their pipeline detects false positive (FP) activations under relevant token injections; and false negatives (FN) under meaning-preserving paraphrases of activating examples. A truly high-level feature should intuitively minimize FP/FN events. We show that Feature Nonlocality, as an independent gradient-based measure, has significant correlations with both FP/FN metrics (Spearman $\rho(\text{FNL},\text{token-injection FP}) = -0.46$, $p<5\times 10^{-4}$; $\rho(\text{FNL}, \text{paraphrase FN}) = 0.27$, $p=0.01$). Unlike those measures, FNL does not rely on LLM judgement or curated contrastive datasets. Together, these results support FNL as an empirical axis of feature abstraction level.

Beyond representational analysis, we demonstrate downstream applications of FNL: (1) We compute the nonlocalities of SAE-filtered, jailbreaking-mitigation features \cite{assogba2026sae}, and found striking evidence that features which successfully undermine wrapper-style jailbreaking attempts often represent surface-level, positional indicator features, rather than truly ``recognizing'' the harmful intents, which provides new insight on the mitigation mechanism; (2) For the model DeepSeek-R1-Distill-Llama-8B, we evaluate steering utility of high-FNL features in improving reasoning performance. We find that steering high-FNL features improves MATH-500 accuracy by $4.6$ points over the unsteered model, whereas low-FNL and random-feature baselines gain $3.8$ and $3.6$ points. As a feature selection criteria, higher-FNL features outperforms lower-FNL features with marginal statistical significance, and the high-FNL arm matches a representative feature selected by the token-cue criterion ReasonScore \cite{galichin2025reasoning}. Note the latter experiment should be treated as a proof-of-concept because we do not generally expect feature nonlocality to reliably predict steering utility \cite{arad2025outputscore}. To conclude, our contributions include

\begin{enumerate}
    \item We introduce \textit{Feature Nonlocality} (FNL), a label-free, gradient-based metric, defined to characterize an SAE feature's typical contextual reach.
    \item We validate FNL as an empirical axis of feature abstraction level through both human judgements and statistical correlations with two proxy measures of abstraction proposed in previous works.
    \item We show that for safety-relevant behaviours, FNL provides useful diagnosis in distinguishing token-level versus abstract features and provides new insights into the mechanism of steerable jailbreaking mitigation.
    \item As a proof of concept, we show that an FNL-selected feature envelope steers the DeepSeek-R1-Distill-Llama-8B model above its unsteered baseline on MATH-500 without relying on labels or token-cue filters.
\end{enumerate}

This work is published under the Agentic Publication Protocol \citep{lu2026agentic}. In the released codebase we have included explicit instructions and technical contexts for an AI coding agent. The reader is welcome to open our repository\footnote{\url{https://github.com/lccqqqqq/sae-feature-nonlocality}} with such an agent, and interact with the agent for explaining technicalities and reproducing results.

\section{Related Work}
In this section, we review the preliminaries of Sparse Autoencoders and activation steering. We then discuss established feature selection and validation approaches in SAE-based interpretability research and note that they do not generally distinguish semantically abstract features from token-level ones.
\paragraph{Sparse Autoencoders (SAEs)} 
SAEs are autoencoders with one hidden layer. Operationally, given an activation $\mathbf{h} \in \mathbb{R}^{d}$, an SAE computes
\begin{equation}
    \mathbf z(\mathbf h) = \sigma\!\left(W_{\mathrm{enc}}\mathbf h+b_{\mathrm{enc}}\right),
    \quad
    \widehat{\mathbf h}= W_{\mathrm{dec}}\mathbf z(\mathbf h)+b_{\mathrm{dec}},
\end{equation}
where $\sigma$ is nonlinearity, $\mathbf z\in\mathbb R^m$ is a sparse latent representation, typically with $m>d$. SAEs are trained to reconstruct hidden activations (typically from a language model) under sparsity constraints. A common choice of sparsity regularizer is the $\ell_{1}$ norm, so that the training objective is schematically
\begin{equation}
    \mathcal{L}_{\mathrm{SAE}} = \left\|\hat{\mathbf{h}} - \mathbf{h}\right\|_{2}^{2} + \alpha \left\|\mathbf{z}(\mathbf{h})\right\|_{1}
\end{equation}
where $\alpha$ is a hyperparameter. Empirically, SAEs are capable of learning useful features for sparse reconstructions that are also monosemantic and human-interpretable. We also note that many latents from SAEs suffer from feature absorption/splitting, and reconstruction-sparsity tradeoffs which adversely affect their interpretability, despite efforts on architectural variants and training setups aiming to mitigate such effects.

\paragraph{Activation Steering} Given a feature direction, \textit{steering} experiments edit the model's internal activations at inference time, by augmenting/suppressing the activation's projection along the feature directions. Common approaches involve \emph{additive steering}~\cite{turner2023activation, rimsky2024steering}; \emph{multiplicative steering}~\cite{zou2023representation}; and \emph{clamping}, which overwrites a feature's activation to a target value~\cite{templeton2024scaling}. Writing $z_a^{\max}$ for a feature's typical peak activation and $\mathbf{e}_a$ for its unit decoder direction,  one may clamp the activation to
\begin{equation}\label{eq:steering}
    \mathbf{h} \;\mapsto\; \mathbf{h} + \big(\gamma\, z_a^{\max} - \mathbf{h}\cdot\mathbf{e}_a\big)\,\mathbf{e}_a ,
\end{equation}
where the gain $\gamma > 1$ sets the steering strength. When there is a subset containing many SAE features of interest, \cite{soo2025fgaa,he2025saessv} construct steering directions from a trained weighted combination of features. Steering experiments validate the causal control power of features in eliciting the behaviour pattern of interest.

\paragraph{Feature Selection/Validation in SAEs}

Unsupervised SAE training produces a task-agnostic dictionary. Therefore, downstream applications require post-hoc criteria for selecting features that are \emph{task-relevant}, \emph{causally effective} and \textit{mechanistically interpretable}.

To ensure task-relevance, \textit{Keyword filtering }looks for features activating on or decoding to task-specific token cues, such as ``wait'', ``but'' for reasoning tasks \cite{galichin2025reasoning}; \textit{Activation-based filtering} selects SAE features based on their activation statistics. \cite{assogba2026sae} finds jailbreaking-relevant features by preparing contrastive datasets of bare harmful requests vs.\ jailbreak harmful prompts inside a wrapper, and selects features with large activation differences over these contrastive prompts; \cite{cho2025corrsteer} directly selects features whose activations correlate with task performance. While useful for identifying feature candidates, such methods do not exclude mechanistically simple, lexical/formatting correlates of the target behaviour \cite{fang2026controllablellmreasoningsparse}. For causal power, Arad et al.\ compute \textit{output score} \cite{arad2025outputscore} diagnosing features' steering utility; other works \cite{galichin2025reasoning, cho2025corrsteer} postselect features by direct validations using steering experiments. 

While feature-filtering and steering validation are well-established pipelines in SAE interpretability works, the mechanistic complexity of these features remains largely unexplored. We note that steerability alone does not establish that a feature implements the high-level computation associated with the steered behaviour. Recent work~\cite{ma2026injection} tests contrastively selected reasoning features and falsifies many as representing lexical correlates rather than genuine reasoning computations, yet the falsification pipeline relies on curated datasets and LLM-generated paraphrases. On the other hand, auto-interpretation methods generate natural-language descriptions of features from their activating contexts and evaluate whether these descriptions predict activation on held-out examples \cite{bills2023language,paulo2025autointerp}. Up to brittleness caveats of LLM-generated descriptions, they offer semantic interpretations of scalable sets of features. However, they do not directly characterize the computation producing their activations. In this work, we propose that Feature Nonlocality provides a complementary, LLM-free measure of a feature's mechanistic complexity, which bears predictive power for distinguishing token-cue features from genuinely high-level features.

\section{Method: Feature Nonlocality}

In this section, we present the definition of Feature Nonlocality and explain the intuition behind the definition. We also give an empirical recipe of its estimation.

\begin{definition}[Feature Nonlocality]
    \label{def:feature-nonlocality}
    Fix an SAE operating on layer $\ell$ of the LLM residual stream. Denote the context window of the LLM by $T$ and consider a prompt $\mathcal{P}$ of $T$ tokens. We use $\mathbf{h}_{T}$ to denote the corresponding $\ell^{\mathrm{th}}$ layer hidden activation at last token position $T$; and $\mathbf{x}_{t}, (1 \leq t \leq T)$ for the token representations at the layer-0 residual stream, after passing through the embedding layer. Let $z^{(\ell)}_{a}(T, \mathcal{P}) := [\sigma(W_{\mathrm{enc}}\mathbf{h}_{T} + b_{\mathrm{enc}})]_{a}$ be the activation of layer-$\ell$ feature $a$ at position $T$ of the prompt $\mathcal{P}$. We omit the layer index $\ell$ to declutter. When the feature is \emph{active} (i.e. $z_{a}(T, \mathcal{P})> \tau$ where $\tau$ is the threshold to assert a firing event), for all $t \leq T$ we can compute the \emph{per-position influence} of $\mathbf{x}_{t}$ on $z_{a}(T, \mathcal{P})$ as the squared gradient norm
    \begin{equation}\label{eq:per-position-influence}
        J_a(t, T, \mathcal{P}) := \left\lVert \frac{\partial z_a(T, \mathcal{P})}{\partial \mathbf{x}_{t}} \right\rVert_2^2 ,
    \end{equation}
    Normalizing Eq.~\eqref{eq:per-position-influence} leads to a valid probability distribution over prefix positions, given a prompt $\mathcal{P}$.
    \begin{equation}\label{eq:normalized-per-position-influence}
        p_{a}^{(\mathcal{P})}(t) := \frac{J_a(t, T, \mathcal{P})}{\sum_{t' \leq T} J_a(t', T, \mathcal{P})}.
    \end{equation}
    The \emph{per-prompt nonlocality} of feature $a$ (relative to the prompt $\mathcal{P}$) is then defined as the entropy of the feature influence distribution.
    \begin{equation}\label{eq:feature-nonlocality}
        H(a, \mathcal{P}) := -\sum_{t=1}^{T} p_{a}^{(\mathcal{P})}(t) \log_2 p_{a}^{(\mathcal{P})}(t)
    \end{equation}
    We also define a dataset-level statistics of the feature nonlocality by averaging Eq.~\eqref{eq:feature-nonlocality} over a set of firing events $\mathcal{D}':= \{\mathcal{P} \in \mathcal{D} | z_{a}(T, \mathcal{P}) > \tau \}$, so that $H(a, \mathcal{D})\equiv \tfrac{1}{|\mathcal{D}'|} \sum_{\mathcal{P} \in \mathcal{D}'} H(a, \mathcal{P})$, where $\mathcal{D}$ represents some dataset of prompts.
\end{definition}

We build some intuition for Definition~\ref{def:feature-nonlocality} with the following remarks. The per-position influence $J_a(t, T, \mathcal{P})$ of Eq.~\eqref{eq:per-position-influence} measures how much the feature's activation at the query position responds to a perturbation of the input embedding at a preceding position $t$; its normalized form $p_a^{(\mathcal{P})}(t)$ is a probability distribution over prefix positions. Such a construction effectively measures the \textit{contextual reach} of a feature. This motivates the hypothesis that topically abstract, context-dependent features have higher FNL than features driven by local lexical or positional cues. The correlation bewteen FNL and semantic abstractness will be discussed later in Sec. \ref{sec:abstractness}.

The notion of FNL is notably independent of SAE features and can be naturally extended to any linear subspace of the hidden activation vector space. In particular, replacing $z_{a}(T, \mathcal{P})$ by $\mathbf{h}_{T}$ itself, we can compute the nonlocality of the residual stream vector directly, which we denote as $H(\mathbf{h}_{T})$.

We note that the operational choices in Eqs.~\eqref{eq:normalized-per-position-influence}-\eqref{eq:feature-nonlocality} are not unique. For example, Eq.~\eqref{eq:normalized-per-position-influence} could be replaced by a softmax over influences, and the Shannon entropy of Eq.~\eqref{eq:feature-nonlocality} could be replaced by a R\'enyi entropy or the inverse participation ratio. We defer a systematic comparison of these choices to future work.

Concretely, fixing a particular feature $a$ of an SAE, we compute its nonlocality with the following steps:
\begin{enumerate}
    \item Run forward passes of prompts sampled from some dataset $\mathcal{D}$ (e.g. GSM8K/WikiText), gather top-$k$ firing events $\mathcal{D}'$ of the feature $a$.
    \item Consider a context of $T$ tokens preceding each firing event (inclusive of the firing token), rerun a backward pass to compute the gradients Eq.~\eqref{eq:per-position-influence}.
    \item Postprocess to obtain for each prompt $\mathcal{P}_i \in \mathcal{D}'$ a per-event entropy $H(a, \mathcal{P}_{i})$ following Eq. \eqref{eq:feature-nonlocality}, then average over all events in $\mathcal{D}'$ to obtain $H(a)$.
\end{enumerate}
For experiments in the main text, we fix $k=32$ and $T=128$ unless stated otherwise. Discussions of the relationship between feature nonlocality and revealed contexts are deferred to the supplementary material. We also note that the computation can be made parallel for all features in the same layer/SAE with one forward pass over the dataset $\mathcal{D}$.

\section{Experiments: Initial Investigations}
\label{sec:experiments}

Before validating the mechanistic functions, we first establish FNL as a stable empirical metric: we examine whether FNL is stable across prompt datasets, how its distribution changes with depth, and whether nearby SAE decoder directions exhibit similar FNL. Throughout this section, we use Gemma-2-2B paired with GemmaScope SAEs \citep{lieberum2024gemmascope} on all-layer residual streams. Results for other model-SAE pairs show consistent findings and are deferred to the supplementary material.

\paragraph{Cross-dataset stability}

We study whether the rank, and absolute values of feature nonlocality are preserved across different corpora. On Gemma-2-2B across several layers, we consider three topically diverse datasets $\mathcal{D} \in \{\text{WikiText}, \text{GSM8K}, \text{Code-Python}\}$~\cite{merity2016pointer,cobbe2021gsm8k,kocetkov2022stack}. For each jointly activated SAE feature $a$ we compute $H(a, \mathcal{D})$. We then compute the Spearman correlation between $H(a, \mathcal{D}_{\text{WikiText}})$ and $H(a, \mathcal{D}_{\text{GSM8K}})$, and similarly for other pairs. Table~\ref{tab:corpus-stability} shows high correlation between feature nonlocalities across different datasets across different layers, and the ranking of features by nonlocality is substantially preserved across corpora.

\begin{table}[t]
\centering
\small
\setlength{\tabcolsep}{2pt}
\begin{tabular}{lccc}
\toprule
Layer & Wiki--GSM8K & Wiki--Code & GSM8K--Code \\
\midrule
\phantom{0}5 & 0.918 & 0.859 & 0.853  \\
12           & 0.815 & 0.826 & 0.766  \\
20           & 0.726 & 0.734 & 0.710  \\
\bottomrule
\end{tabular}
\caption{Cross-corpus stability of feature nonlocality in Gemma-2-2B. Each cell ranks $2{,}000$ features independently on WikiText, GSM8k and Code-Python. Here we use $k=60$ firing events per feature per corpus and context window $T=128$. Each entry reports the Spearman agreement of per-feature $H(a, \mathcal{D})$ between pairs of corpora. The ranking is largely preserved at every depth, although the Spearman correlation fall slightly with depth.}
\label{tab:corpus-stability}
\end{table}

\begin{figure}[t]
    \centering
    \includegraphics[width=\columnwidth]{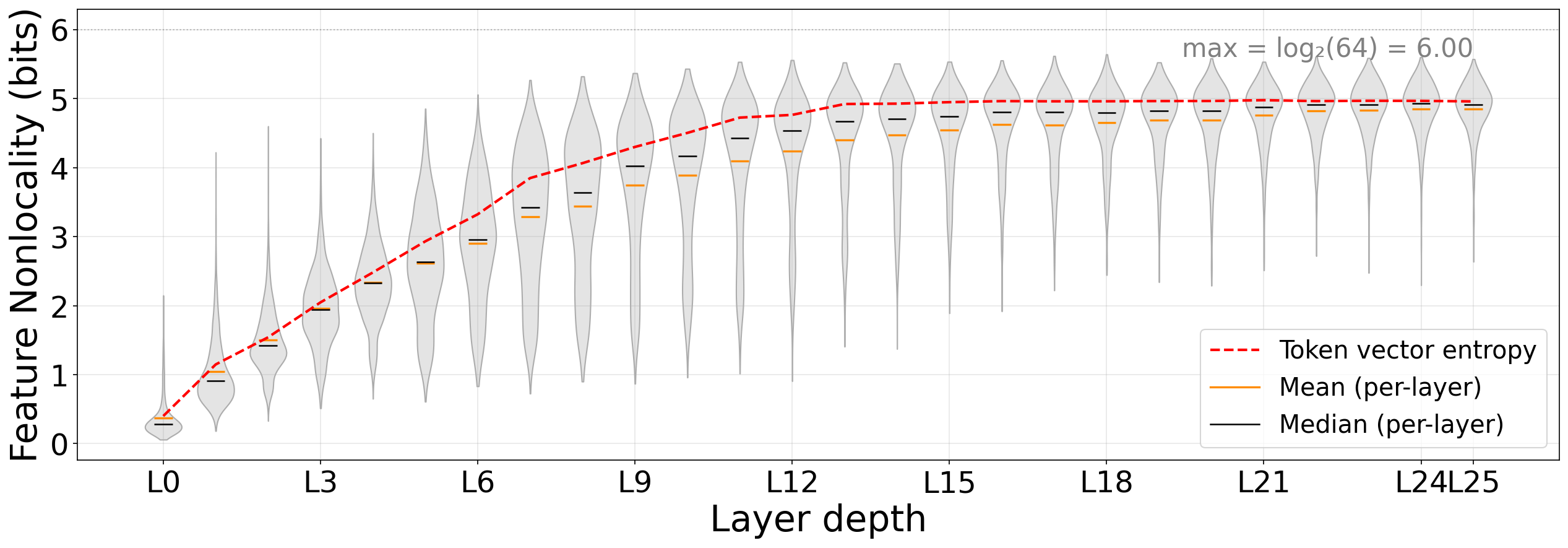}
    \caption{GemmaScope feature nonlocality as a function of depth for Gemma-2-2B on WikiText (context window $T = 64$). Each violin visualizes the distribution of empirically estimated $H(a)$ at the corresponding layer, overlaid with the per-layer mean (orange) and median (black), as well as the residual-stream baseline $H(\mathbf{h})$ (red dashed). The mean nonlocality saturates by the middle layers ($\ell \approx 13$--$16$).}
    \label{fig:nonlocality-depth}
\end{figure}

\paragraph{Dependence with layers}
For Gemma-2-2B and GemmaScope SAEs, we feed a subset of WikiText and extract FNL statistics for layers $\ell = 0, 1,..., 25$. Progressing deeper into the network, FNL shifts upward and plateaus at around the middle layers (Fig. \ref{fig:nonlocality-depth}), broadly following the corresponding residual-stream statistic $H(\mathbf{h}_{T}^{(\ell)})$. Testing on other model-SAE pairs over a range of 0.5B-9B models in Gemma, Llama, Qwen families; as well as varying the datasets $\mathcal{D}= \{\text{WikiText}, \text{GSM8K}, \text{Code-Python}\}$ demonstrates robustness of the rise-then-saturate pattern. This depth-dependent trend is suggestive, although not conclusive, evidence that later layers tend to process more contextual, semantic information \cite{jawahar2019bertfolklore}.

\paragraph{Feature nonlocality and Feature Geometry}

\begin{figure}[t]
    \centering
    \includegraphics[width=\columnwidth]{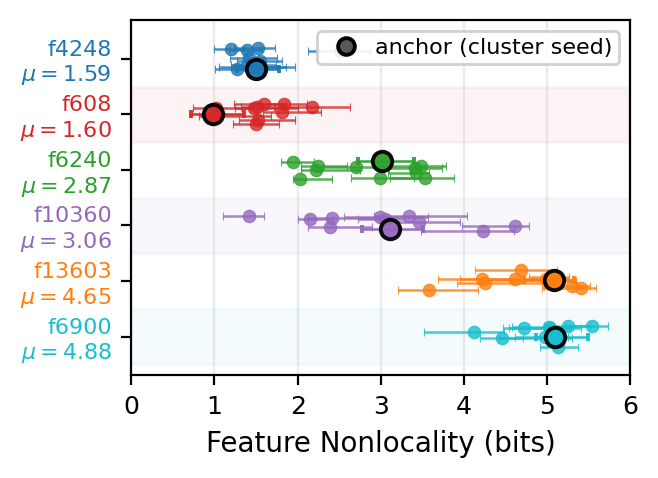}
    \caption{Decoder-cosine geometry pins $H(a)$ (Gemma-2-2B layer~12). For each of six anchor features we take its top-10 decoder-cosine neighbours and plot every feature's $H(a)$. For each individual feature, the computation of FNL samples top-$k$ ($k = 32$) max-activation firing events across a sample WikiText corpus; the error bar represents the IQR of FNL statistics.}
    \label{fig:feature-clusters-2d}
\end{figure}

In Fig.~\ref{fig:feature-clusters-2d} we focus on layer-12 and show that FNL varies smoothly with local decoder geometry. Across six anchor features spanning the observed FNL range, their nearest neighbours under decoder cosine similarity $\mathrm{sim}(a,b):= \cos \langle \mathbf{e}_{a}, \mathbf{e}_{b}\rangle$ tend to occupy similar ranges of FNL. 
Because decoder cosine characterizes the direction in which a feature writes to the residual stream, we interpret this result as a consistent association between FNL as a read-side contextual reach measure and write-side feature geometry. This supports treating Feature Nonlocality as a stable metric correlating with features' write-side mechanistic functionalities.

More speculatively, since FNL can be defined for arbitrary linear subspaces, such correlations between contextual reach and decoder geometry are suggestive of an appealing picture that the residual-stream space may decompose into directions of different contextual reach, and the SAE features form a sparse overcomplete basis of this organization. We leave related investigations on residual stream geometry to future work.

\section{Nonlocality and Semantic Abstractness}
\label{sec:abstractness}

\begin{figure*}[t]
    \centering
    \includegraphics[width=\textwidth]{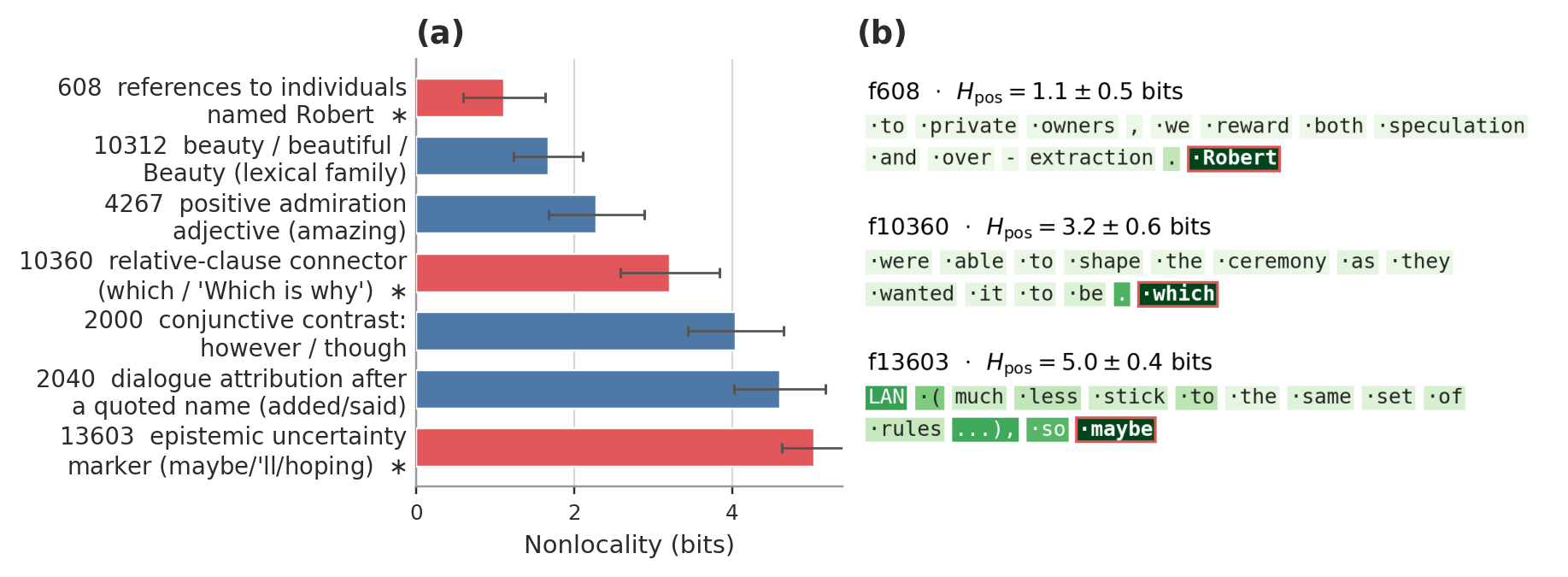}
    \caption{Feature nonlocality traces a concrete-to-abstract continuum (Gemma-2-2B layer~12, $T = 128$). \textbf{(a)} Seven representative features ranked by mean $H(a)$ ($\pm$ std across activating events); the three marked ($\ast$, red) are the features shown in panel (b). Single-token detectors (e.g.\ f608, ``Robert'') sit at the low-nonlocality end and utterance-level features (e.g.\ f13603, epistemic hedging) at the high end. \textbf{(b)} Per-position influence $J_a(t')$ over the top-three max-activating contexts of those three features; each token's background is shaded by $|J_a(t')|$ and the activating token is boxed. Influence is concentrated on a single token for the low-$H(a)$ feature f608 and grows progressively more diffuse for the mid-$H(a)$ relative-clause connector f10360 ($3.2$ bits) and the high-$H(a)$ epistemic-uncertainty feature f13603 ($5.0$ bits).}
    \label{fig:feature-continuum}
\end{figure*}

In this section, we present evidence that feature nonlocality $H(a)$ correlates with existing proxy, falsification-based measures of semantic abstractness \cite{ma2026injection}, distinguishing token-level features from high-level concept features. To start, Fig.~\ref{fig:feature-continuum} presents representative feature samples from Gemma-2-2B layer 12, ranked by $H(a)$. Following the auto-interp description on the $y$-axis, we observe the apparent shift from lexical (e.g. f608 ``Robert indicator'') to contextual (e.g. f13603 ``uncertainty marker'') with increasing nonlocality. We now make this correlation quantitative.

\paragraph{Token injection susceptibility} Genuine high-level features should \emph{not} be switched on when only the trigger token is present but not the associated semantics. We conduct token injection experiments over 100 features that fire systematically more frequently in reasoning traces than non-reasoning text, selected in DeepSeek-R1-Distill-Llama-8B. We inherit the methodology from \cite{ma2026injection}, which we summarize below for completeness. For each feature we take its top-activating unigrams, bigrams and trigrams, splice them into non-reasoning passages while preserving coherence. We compare the feature's activation over injected passages against unmodified passages. With a truly contextual feature, we expect the activation statistics to remain unmoved by this manipulation. Let the target feature activations for injected/original passages be $z_{\mathrm{inj}}$ and $z_{\mathrm{orig}}$, respectively, we measure the \textit{activation recovery} by $d:= \frac{(\langle z_{\mathrm{inj}} \rangle - \langle z_{\mathrm{orig}} \rangle)}{\sqrt{(\mathrm{Var}(z_{\mathrm{inj}}) + \mathrm{Var}(z_{\mathrm{orig}})) / 2}}$ \cite{cohen1988d}. This then classifies features as token-driven ($d \ge 0.8$), partially token-driven ($0.5 \le d < 0.8$), weakly token-driven ($0.2 \le d < 0.5$), and context-dependent ($d < 0.2$, or not significant). Strong injection recovery trivializes the feature, in that lexical cues alone suffice to elicit it. At layer-19, we reproduced the protocol in \cite{ma2026injection} and independently compute the nonlocalities Eq. \eqref{eq:feature-nonlocality} of the same set of features. In Table \ref{tab:injection-layers}, we find FNL correlates negatively with injection recovery ($\rho = -0.39$ to $-0.46$) and consistently discriminates token-driven (TD) from context-dependent (CD) features (AUC 0.73-0.84). Therefore, higher FNL implies lower susceptibility to token-cue elicitation.

\paragraph{Paraphrase Invariance} 
Token injection tests for false-positive activations. A complementary validation, which aims to test for false-negatives, relies on the intuition that genuine high-level features should retain the activation under paraphrases that preserve the semantic information. We again adapt the pipeline in \cite{ma2026injection}. Given an SAE feature $a$, we collect its firing contexts, and for each firing we compare (1) the original activation $z_{\mathrm{orig}}$; (2) a length-calibrated paraphrase intended to preserve meaning $z_{\mathrm{para}}$; (3) a token shuffle that preserves the tokens while destroying the meaning $ z_{\mathrm{shuffle}}$; (4) activation baseline of a random, unrelated prompt $z_{\mathrm{base}}$. Define \textit{retention} as $R_{a}(z_{\mathrm{para}}) := \frac{z_{\mathrm{para}}- z_{\mathrm{base}}}{ z_{\mathrm{orig}} - z_{\mathrm{base}}}$ and similarly for $R_{a}(z_{\mathrm{shuffle}})$, characterizing the (properly baselined) fraction of retained activation. The \textit{paraphrase robustness} score for the feature is then defined by $S_{a} = \mathrm{med}(R_{a}(z_{\mathrm{para}})) - \mathrm{med} (R_{a}(z_{\mathrm{shuffle}}))$, where $\mathrm{med}$ takes the median from the statistics of all sampled firing events. Positive values for $S_{a}$ imply greater robustness to paraphrasing than to order destruction, while negative $S_{a}$ falsifies the claim that $a$ is semantically abstract. 

\begin{figure}[t]
    \centering
    \includegraphics[width=\columnwidth]{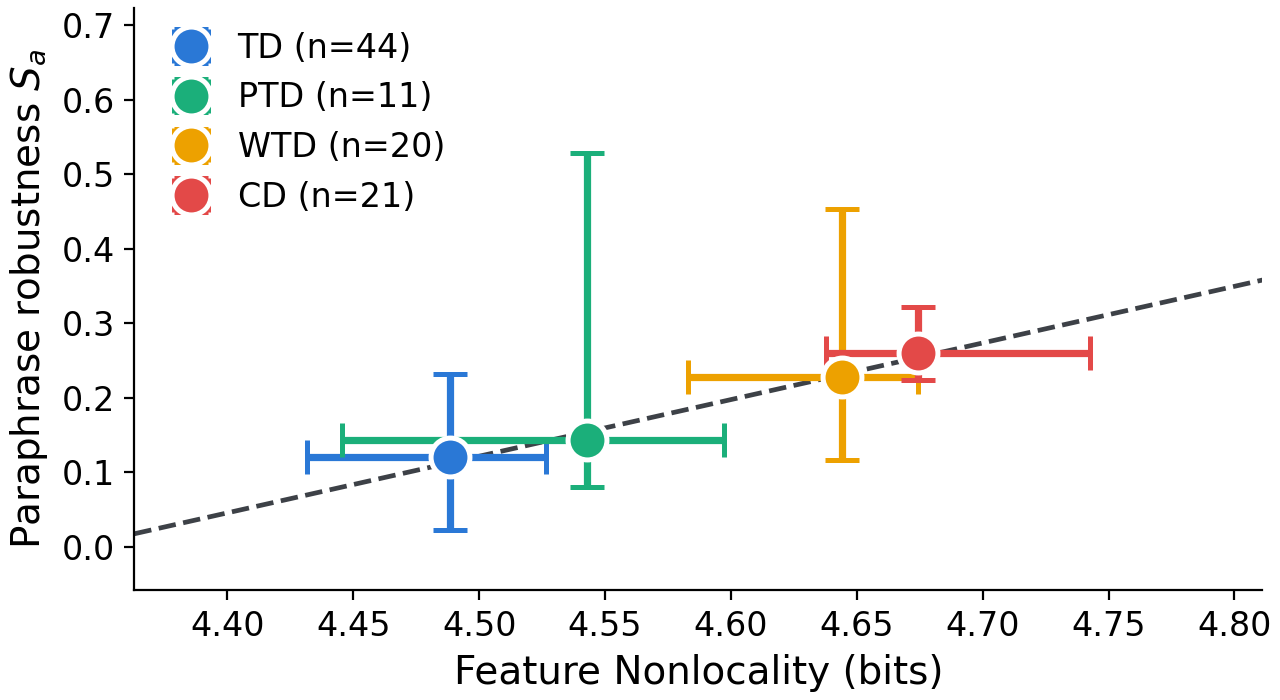}
    \caption{Paraphrase robustness $S_a$ versus nonlocality, presentation collapsed to the four token-injection classes (median $\pm$ 68\% bootstrap CI on both axes, dashed line fitting through the class medians for readability). Per-feature Spearman correlation, $\rho = +0.27$ ($p = 0.011$, $n = 96$ features).}
    \label{fig:paraphrase-binned}
\end{figure}

We again identify positive correlation of the paraphrase robustness $S_{a}$ with the nonlocality of feature $a$ ($\rho=0.27$, $p=0.011$), as illustrated in Fig. \ref{fig:paraphrase-binned}. We also note that the median of $S_{a}$ increases monotonically across the TD-CD spectrum, providing independent cross-validation between lexical susceptibility and paraphrase robustness.

These two tests probe semantic abstractness from complementary directions. Intuitively, a feature's semantic abstractness anticorrelates with its activation susceptibility to token injection, whereas it correlates with robustness under meaning-preserving paraphrases. On both axes we report that feature nonlocality, as an independent gradient-based metric, reproduces those correlations. Up to modest fluctuations, FNL also successfully distinguishes token-level vs.\ context-dependent features. We therefore conclude that FNL, as a measure of contextual dependence of a feature, provides an LLM-independent correlational measure of a feature's semantic abstraction level.

\begin{table}[t]
\centering
\begin{tabular}{lcccc}
\toprule
Layer & Dictionary & $\rho$ & AUC & TD / CD mean \\
\midrule
10 & LlamaScope 32k & $-0.41$ & 0.81 & 4.03 / 4.57 \\
11 & LlamaScope 32k & $-0.42$ & 0.77 & 4.13 / 4.59 \\
12 & LlamaScope 32k & $-0.46$ & 0.84 & 4.19 / 4.71 \\
19 & Galichin 65k & $-0.39$ & 0.73 & 4.38 / 4.68 \\
\bottomrule
\end{tabular}
\caption{On DeepSeek-R1-Distill-Llama-8B, across four layers and two SAE dictionaries~\cite{he2024llamascope,galichin2025reasoning}, we report the Spearman correlation $\rho$ between Feature Nonlocality and the activation recovery under token-injection; AUC measures how well FNL separates token-driven (TD) from context-dependent (CD) features, and is equivalently the probability that a randomly drawn CD feature has higher FNL than a randomly drawn TD feature, with $0.5$ indicating chance; the final column gives mean FNL in bits for the two classes.}
\label{tab:injection-layers}
\end{table}

\section{Downstream Applications}
\label{sec:applications}

We demonstrate the use of Feature Nonlocality with two case studies of downstream tasks. First, we audit the mechanism for a class of jailbreaking-mitigation features, previously detected via contrastive activation enrichment \cite{assogba2026sae}. We demonstrate that FNL bears \textit{diagnostic} value in resolving mechanistic functionalities. Second, we conduct steering experiments on high-FNL features and find improvements on reasoning benchmarks at least comparable with a representative feature selected by token-cue filtering \cite{galichin2025reasoning}. This suggests that FNL offers useful \textit{selection} value for effective interventions of high-level behaviours.

\subsection{Auditing Jailbreaking-mitigation mechanisms}
\label{sec:ccdelta-audit}

CC-Delta~\citep{assogba2026sae} filters SAE features that defend against wrapper-based jailbreaking attempts. Specifically, they construct contrastive plain harmful prompts $\mathcal{P}$ vs.\ the same prompts wrapped for jailbreaking $\mathcal{P}_\mathrm{wrapper}\mathcal{P}$. Then the authors select features with drastically different activations on the harmful prompt tokens of the contrastive datasets. They report that steering on the derived features improves jailbreak defense against out-of-distribution, Few-Shot-Json attacks (a type of wrapper attacks, embedding harmful requests inside json-formatted benign question-answer pairs).

After reproducing their feature selection pipelines on the same model (DeepSeek-R1-Distill-Llama-8B) and SAE (LlamaScope \cite{he2024llamascope}, layer-17 residual stream), we further compute their nonlocality via backward passes over the same dataset. Strikingly, we found that 21 of the 25 selected features have 0 nonlocality. They activate almost exclusively at position-0 of the plain harmful prompt (but not on the same prompt tokens when they are wrapped), giving zero contextual reach by definition. The remaining 4 features are content-aware and exhibit nontrivial FNL. Steering over the positional/attention-sink-like features \cite{xiao2024streaming} vs.\ content-aware features separately, we find only the positional features lead to substantial improvements for OOD jailbreaking mitigation, whereas steering on the content feature subset stays near the unsteered baseline (Table \ref{tab:ccdelta-decompose}).

Through the lens of Feature Nonlocality, we may sharpen the mechanism of jailbreak-mitigating features as follows. While successfully eliciting OOD jailbreak defense, these effective features are typically positional and have low FNL. Thus the features should be understood as surface-level, beginning-of-sequence (BOS) indicators rather than encoding the computation for recognizing harmful intents. A robust mechanistic account of these features is left to future work.

\begin{table}[t]
\centering
\begin{tabular}{lccc}
\toprule
Component & $\lVert v \rVert$ & held-out & plain \\
\midrule
unsteered baseline & 0 & 0.511 & 0.963 \\
all 25 features & 3.19 & 0.887 & 0.963 \\
positional subset (21) & 2.04 & \textbf{0.911} & 0.915 \\
content subset (4) & 2.40 & 0.504 & 0.954 \\
random dir.\ ($\lVert v \rVert{=}2.04$) & 2.04 & 0.630 & 0.943 \\
random dir.\ ($\lVert v \rVert{=}3.19$) & 3.19 & 0.775 & 0.966 \\
\bottomrule
\end{tabular}
\caption{Steering utility of CC-Delta-selected features. The 25 features split into a size-21 positional subset and a size-4 contextual subset. We steer each subset with per-group optimized steering strengths $\lVert v\rVert$. The ``held-out'' column reports safety under the Few-Shot-Json wrapper attack and the ``plain'' column safety on the unwrapped harmful requests, both measuring successful defense rate over the same $404$ StrongReject requests~\cite{souly2024strongreject} The positional subset accounts for most of the improvement.}
\label{tab:ccdelta-decompose}
\end{table}

\subsection{FNL-Guided Steering}
\label{sec:steering-application}

As a supplementary study, we further conduct steering experiments using feature nonlocality as a selection criterion, to see whether intervening on high-FNL features positively affects the model's reasoning capabilities. We steer DeepSeek-R1-Distill-Llama-8B at layer 19 and paired residual-stream SAEs \cite{galichin2025reasoning} with 65,536 features per layer. Feature nonlocality values are computed with the dataset OpenThoughts-114k \cite{openthoughts2025}. In the steering experiment, we select the top 20\% features ranked by their feature nonlocalities, and steer by simultaneously clamping all features in the subset with strength $\gamma$ c.f. Eq.~\eqref{eq:steering}. To find the steering strength, we perform pilot sweeps over steering strengths $\gamma \in [1.0, 1.2]$ and postselect the best-performing $\gamma$ for a held-out subset of MATH-500~\cite{lightman2023letsverify,hendrycks2021math}. For benchmark evaluations, we collect steered model rollouts on MATH-500, which are subsequently graded by a local Llama-3.3-70B judge and report the average score over 4 rollouts.

We consider various baseline steering approaches, including (1) \textbf{Unsteered Baseline}; (2) \textbf{Low nonlocality}: Steering the \textit{bottom} 20\% features ranked by feature nonlocality; (3) \textbf{Random}: Steering a random subset of features, with the same size as the high/low nonlocality arms; (4) \textbf{Representative single feature}: Steering one feature individually at strength $\gamma=2$ selected by ReasonScore \cite{galichin2025reasoning}, which filters for reasoning features via token-level cues.

The results are summarized in Table~\ref{tab:envelope}. All steered models beat the baseline by a nontrivial margin, and the high-FNL envelope is the best-scoring arm while also producing the shortest thinking traces among the envelope arms. Note that our feature selection criterion is based entirely on nonlocality, with no token-level or activation-based filters.

The supplementary material extends these experiments to layers 10--12 of the same model and to Gemma-2-9B and Qwen3-8B. Steering high-FNL features generically beats steering low-FNL ones, but outside DeepSeek-R1-Distill-Llama-8B both arms fall below the unsteered and random-feature baselines. We therefore present the experiment as a proof of concept rather than evidence that FNL reliably predicts steering utility.

\begin{table}[t]
    \centering
    \setlength{\tabcolsep}{4pt}
    \begin{tabular}{lccc}
        \toprule
        Setting & avg@4 & tokens & $\Delta$\,avg@4 \\
        \midrule
        Unsteered Baseline      & 0.865 & 4{,}214 & --- \\
        High Feature Nonlocality  & \textbf{0.911} & 3{,}918 & $+4.6$ \\
        Low Feature Nonlocality   & 0.903 & 4{,}095 & $+3.8$ \\
        Random        & 0.901 & 3{,}944 & $+3.6$ \\
        \midrule
        f3466 (representative) & 0.904 & 3{,}894 & $+3.9$ \\
        \bottomrule
    \end{tabular}
    \caption{Steered model performance evaluated on MATH-500, graded by a local Llama-3.3-70B judge; entries are avg@4 (mean accuracy over four rollouts). The last row steers a single representative feature, f3466, individually at strength $\gamma=2$ following~\cite{galichin2025reasoning}. We observe that collectively steering high-FNL features outperforms representative features selected via curated token-level filters.}
    \label{tab:envelope}
\end{table}

\section{Conclusion}

We introduced feature nonlocality (FNL), the entropy of an SAE feature's per-position influence over the causal prefix, as a label-free measure of how much context a feature actually reads. The measure requires no contrastive datasets or interventions, yet it is demonstrated to quantitatively correlate with the semantic abstractness of features.

Such a quantity is helpful for pursuing a fine-grained mechanistic understanding of LLMs' emergent, abstract behaviours, now increasingly perceived as an urgent task in AI safety. The distinction we advocate is between what an intervention achieves and what the intervened feature represents: the former is validated by steering utility as in various interpretability works, while the latter requires additional evidence which we supply in this paper. Applied downstream, we used FNL to audit the mechanism of jailbreak defense, revealing that its selected features read a positional artifact rather than harmful content; and, as a proof of concept, selected features for envelope steering that improve reasoning-benchmark accuracy over the unsteered baseline on a reasoning-finetuned model without any supervision.

\clearpage
\section*{Data and Code Availability}

The data and codes associated with this work are published in the repository \url{https://github.com/lccqqqqq/sae-feature-nonlocality} (release \texttt{v1.0.0}), which is also the Agentic Publication Protocol (APP) paper repository, following the specifications in \citep{lu2026agentic}. All figures and tables in this work are reproducible in CPU minutes. The codes for generating the raw data for each experiment are also included.

\appendix
\setcounter{table}{0}
\setcounter{figure}{0}
\renewcommand{\thetable}{S\arabic{table}}
\renewcommand{\thefigure}{S\arabic{figure}}
\section*{Supplementary Material}

\section{Feature Nonlocality as a Feature-level Measure}
\label{sec:supp-measurement}

This section reports supplementary experiments on nonlocality as a feature-level measure. We present additional results on the cross-corpus stability of FNL across a wider range of open-source models and layers, and on the depth-dependence of FNL. We also study how FNL depends on the revealed context, i.e. how feature nonlocalities change as we gradually reveal more of it. Finally, in the context of reasoning features, we present a study of how FNL correlates with other established measures obtained from keyword filtering \cite{galichin2025reasoning} or activation differences~\cite{venhoff2025steering,assogba2026sae}.

\subsection{Supplemental Experiments for Cross-dataset stability of Feature Nonlocalities}
\label{sec:supp-corpus}

Table~\ref{tab:corpus-stability} of the main text reports that per-feature nonlocality rankings agree across topically dissimilar corpora in Gemma-2-2B. We extend that measurement to Llama-3-8B and Qwen3-8B, at three depths each. We note that the raw Spearman $\rho$ in Table \ref{tab:supp-corpus} is computed between the \textit{mean} Feature Nonlocality given a set of firing events within their respective corpora. To account for fluctuations of the mean, we consider a baseline, characterizing the ceiling of attainable Spearman correlation as follow: within a single corpus we split each feature's firing events into two disjoint halves and correlate the rankings computed on each half, giving the split-half reliability $r_{\mathcal{D}}$; for a pair of corpora the agreement attainable by a perfectly corpus-invariant measure is $\sqrt{r_{\mathcal{D}_1} r_{\mathcal{D}_2}}$ \cite{spearman1904proof}. We report the disattenuated cross-dataset agreement in the last column of Table \ref{tab:supp-corpus}.

\begin{table}[t]
\centering
\small
\setlength{\tabcolsep}{4pt}
\begin{tabular}{lccccc}
\toprule
model & layer & rel.\ depth & $\rho_{\mathrm{raw}}$ & $\rho_{\mathrm{ceiling}}$ & $\tilde{\rho}$ \\
\midrule
Gemma-2-2B  & \phantom{0}5 & 0.20 & 0.877 & 0.981 & 0.894 \\
Gemma-2-2B  & 12 & 0.48 & 0.802 & 0.954 & 0.841 \\
Gemma-2-2B  & 20 & 0.80 & 0.723 & 0.890 & 0.812 \\
\midrule
Llama-3-8B  & \phantom{0}6 & 0.19 & 0.606 & 0.900 & 0.673 \\
Llama-3-8B  & 15 & 0.48 & 0.404 & 0.761 & 0.531 \\
Llama-3-8B  & 24 & 0.77 & 0.441 & 0.787 & 0.560 \\
\midrule
Qwen3-8B    & \phantom{0}7 & 0.20 & 0.773 & 0.963 & 0.803 \\
Qwen3-8B    & 18 & 0.51 & 0.710 & 0.915 & 0.776 \\
Qwen3-8B    & 28 & 0.80 & 0.669 & 0.887 & 0.754 \\
\bottomrule
\end{tabular}
\caption{Cross-corpus stability of the nonlocality ranking, measured against its own reliability ceiling. The column $\rho_{\mathrm{raw}}$ is the Spearman correlation of per-feature $H(a, \mathcal{D})$ averaged over the three corpus pairs drawn from WikiText, GSM8K and Code-Python \cite{merity2016pointer,cobbe2021gsm8k,kocetkov2022stack}. $\rho_{\mathrm{ceiling}}$ is the agreement attainable if the measure were perfectly corpus-invariant, estimated from split-half reliability within each corpus. $\tilde{\rho} := \rho_{\mathrm{raw}} / \rho_{\mathrm{ceiling}}$ is the disattenuated correlation, which we report as a properly baselined cross-dataset agreement metric. Each cell ranks the same $2{,}000$ features, sampled once from those carrying enough firing events on all three corpora, on up to $k=60$ events per corpus with context window $T=128$. Dictionaries used are GemmaScope \cite{lieberum2024gemmascope}, Llama Scope \cite{he2024llamascope} and Qwen-Scope \cite{deng2026qwenscope}.}
\label{tab:supp-corpus}
\end{table}

Measured against this ceiling, the ranking retains the majority of its attainable agreement at every model and depth, extending the cross-corpus stability of Table~\ref{tab:corpus-stability} beyond the model family of the main text.

\subsection{Depth profile across models and corpora}
\label{sec:supp-depth}

\begin{figure*}[!tp]
    \centering
    \includegraphics[width=\textwidth]{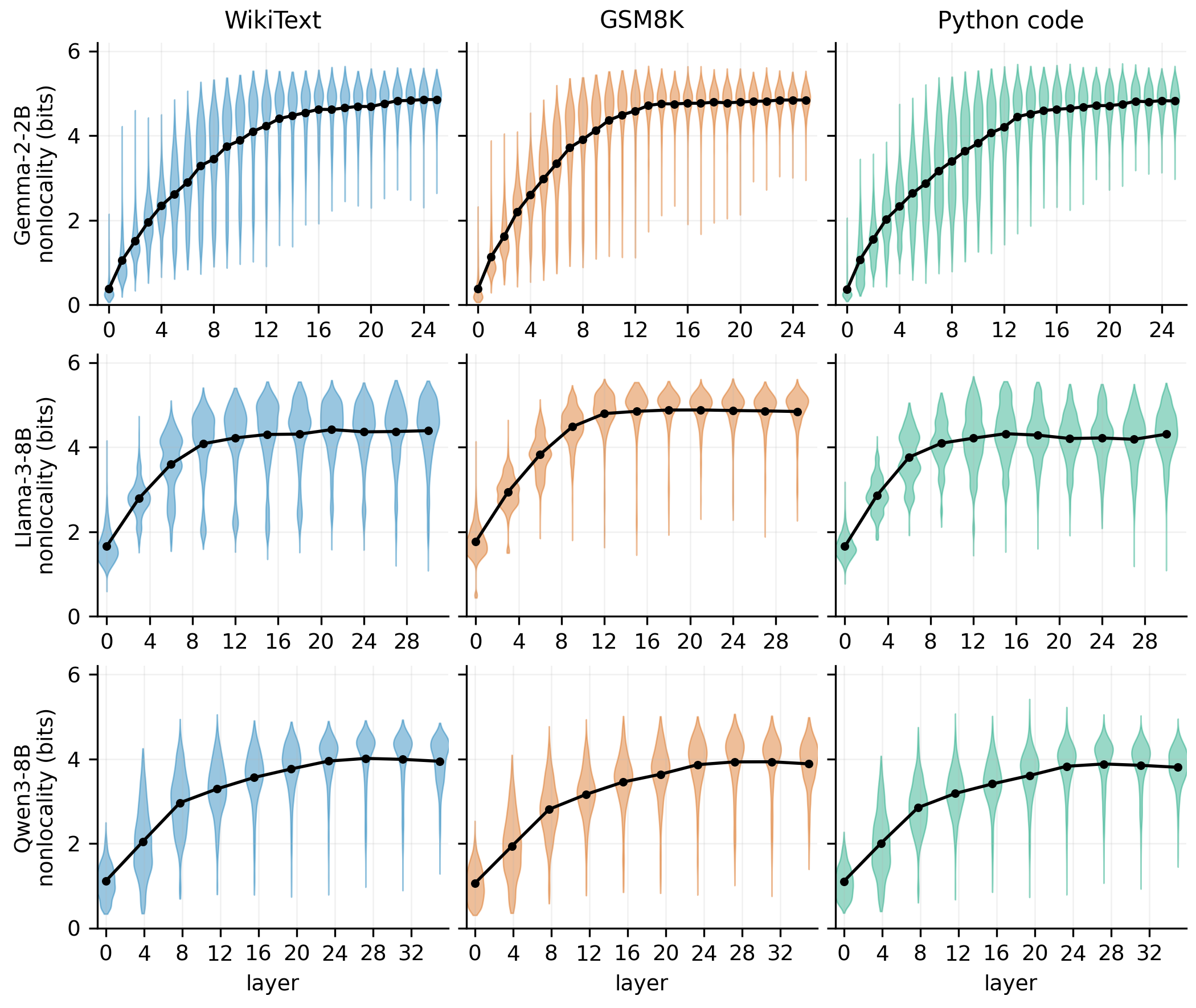}
    \caption{Nonlocality against depth for three models (rows) measured independently on three corpora (columns). Each violin is the distribution of per-feature mean nonlocality at one layer; the black line traces the per-layer mean. Within every row the three profiles nearly coincide,  while the residual separation is a small systematic shift by corpus, GSM8K highest and Python code lowest. The horizontal axis spans each model's true layer range. They uniformly show the same rise-then-saturate shape. }
    \label{fig:fnl-depth-corpus}
\end{figure*}

Figure~\ref{fig:fnl-depth-corpus} repeats the depth sweep behind Fig.~\ref{fig:nonlocality-depth} of the main text on three corpora for the three largest models of our pool: in every panel nonlocality rises over the first half of the network and then saturates, with the three corpus profiles nearly coincident, so the rise-then-saturate shape is an intrinsic property of FNL rather than of any one model, dictionary, or input distribution.

\subsection{Dependence on the revealed context length}
\label{sec:supp-ctxlen}

\begin{figure*}[!tp]
    \centering
    \includegraphics[width=\textwidth]{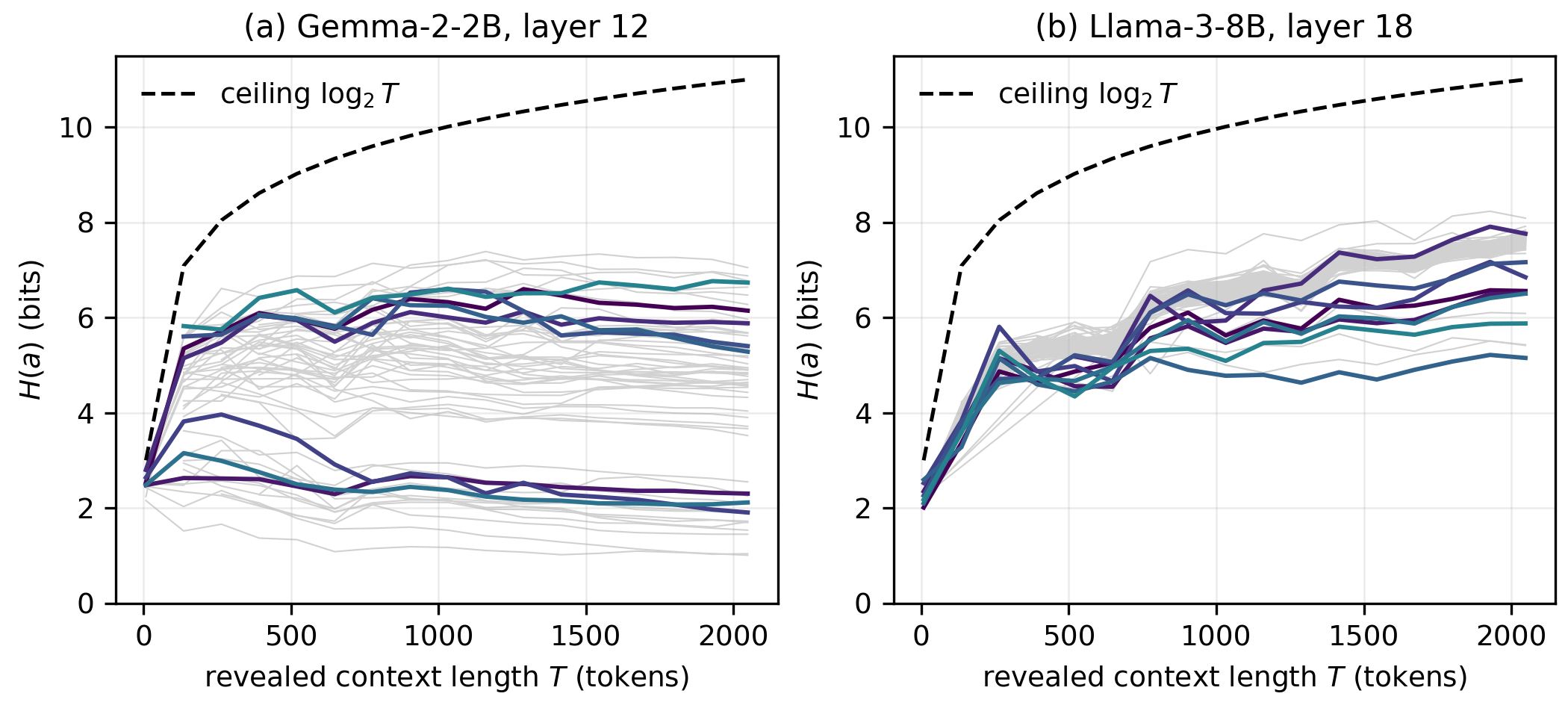}
    \caption{Nonlocality against the revealed context length $T$, on one coherent FineWeb-Edu document, at one mid-depth layer per model. Coloured curves are the eight strongest-firing features, grey curves every other feature active across the sweep, and the dashed curve the uniform-influence ceiling $\log_2 T$. \textbf{(a)} Gemma-2-2B layer 12; \textbf{(b)} Llama-3-8B layer 18. Features separate by several bits and hold their values as the window grows, settling at intrinsic reaches well below the ceiling.}
    \label{fig:fnl-ctxlen}
\end{figure*}

Feature nonlocality naturally depends on how much context is revealed in the feature's ``past lightcone''. To measure this dependence we sweep the revealed window $T$ from a few tokens to a few thousand on a single coherent FineWeb-Edu document \cite{lozhkov2024fineweb}, recomputing every active feature's nonlocality at each $T$, for all layers of Gemma-2-2B and 11 layers of Llama-3-8B; Figure~\ref{fig:fnl-ctxlen} shows every such feature at one mid-depth layer of each model. In both models nonlocality first rises with the revealed context and then saturates. Comparisons at a fixed window, which are the only comparisons the main text makes, are thus sound for Gemma-2-2B and Llama-3-8B, and are expected to generalize to other models.

\subsection{Nonlocality tracks decoder geometry rather than semantic similarity}
\label{sec:supp-geometry}

\begin{figure*}[!tp]
    \centering
    \includegraphics[width=\textwidth]{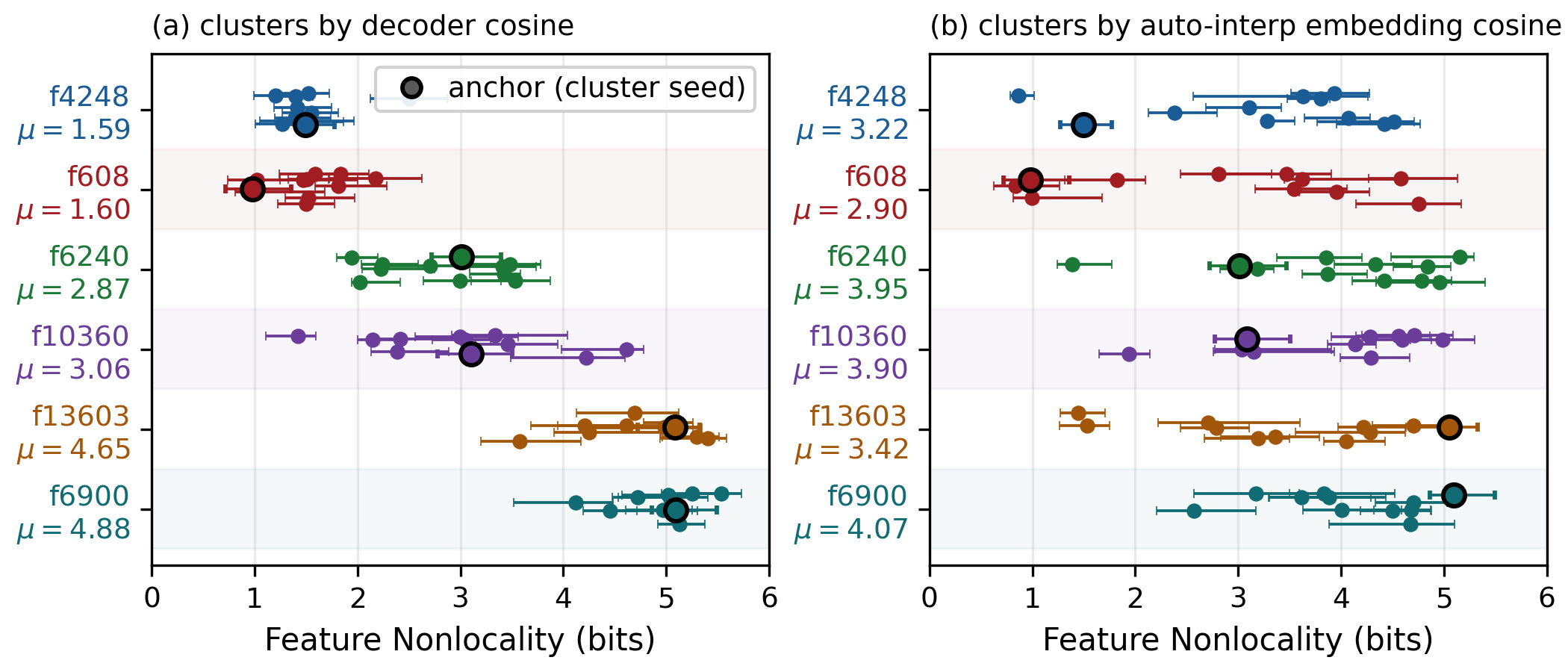}
    \caption{The six anchor features of the main text's Fig.~\ref{fig:feature-clusters-2d} under two notions of similarity, drawn in the format of that figure. \textbf{(a)} Each anchor's ten nearest neighbours by decoder cosine, reproducing the main text's figure: dots are per-feature median nonlocality with IQR whiskers over the feature's firing events, the black-ringed dot is the anchor, and the row label gives the cluster mean $\mu$. This is an exact recapitulation of Figure 2 in the main text. \textbf{(b)} The same anchors with their ten nearest neighbours by cosine similarity between sentence embeddings of the auto-interpretation descriptions, on the same rows and the same axis; embeddings are computed with the BGE-large-en-v1.5 text-embedding model~\cite{xiao2023cpack}. The three nearest members of every cluster under each notion are listed in Table~\ref{tab:supp-clusters}.}
    \label{fig:fnl-clusters-semantic}
\end{figure*}

Figure~\ref{fig:feature-clusters-2d} of the main text shows that features close together under decoder cosine similarity occupy similar ranges of nonlocality. We consider a semantically more direct measure of similarity, from the auto-interpretation descriptions~\cite{bills2023language,paulo2025autointerp}. Each description is embedded with the BGE-large-en-v1.5 text-embedding model~\cite{xiao2023cpack}, and semantic similarity is the cosine between the embeddings. Figure~\ref{fig:fnl-clusters-semantic} draws the six anchor features of the main text's figure under both notions in the same format, and Table~\ref{tab:supp-clusters} lists the nearest neighbours' descriptions. Under decoder cosine (panel a) each cluster occupies a narrow band of the nonlocality axis, although its members' descriptions share might not be obviously related. Under description embedding cosine (panel b) the neighbourhoods of the same anchors are topically uniform almost to the word, yet each spreads across most of the axis. Thus, we conclude that the decoder cosine similarity, as a metric defined on the write-side, correlates better with nonlocality than semantic similarity. In retrospect, this result makes sense because semantically similar concepts may require the feature to attend to different ranges of text. For example, two features may both be described as concerning ``beauty'', while one fires on the word itself, a lexical read with no contextual reach, and the other recognizes a described scene as beautiful, a judgement that requires reading a span of context.

\begin{table*}[!tp]
\centering
\small
\setlength{\tabcolsep}{5pt}
\begin{tabular}{lllp{0.58\textwidth}}
\toprule
similarity & feature & $H(a)$ median [IQR] (bits) & auto-interpretation description \\
\midrule
\multicolumn{4}{l}{\textbf{f4248} (anchor, $H(a)=1.50$ bits): \textit{phrases that signify quality, reliability, or suitability}}\\
decoder & f2054 & $1.55$ $[1.27, 1.81]$ & positive affirmations and expressions of approval\\
 & f2829 & $1.40$ $[1.18, 1.57]$ & contexts where improvement, enhancement, or optimization is being discussed\\
 & f411 & $1.53$ $[1.40, 1.86]$ & references to the concept of ``best'' or ``optimal'' in various contexts\\
\cmidrule(lr){2-4}
semantic & f1820 & $3.93$ $[3.51, 4.28]$ & phrases related to assurance, quality control, and meeting standards\\
 & f6267 & $4.07$ $[3.64, 4.28]$ & phrases that describe suitability and appropriateness in various contexts\\
 & f2431 & $4.52$ $[3.77, 4.71]$ & phrases related to product selection and suitability\\
\midrule
\multicolumn{4}{l}{\textbf{f608} (anchor, $H(a)=0.98$ bits): \textit{references to individuals named Robert}}\\
decoder & f15176 & $1.81$ $[1.59, 2.29]$ & terms and phrases related to smoking and its impact on health\\
 & f5430 & $1.00$ $[0.81, 1.68]$ & mentions of the name ``Robert.''\\
 & f11410 & $1.50$ $[1.22, 1.78]$ & instances of the word ``once''\\
\cmidrule(lr){2-4}
semantic & f5430 & $1.00$ $[0.81, 1.68]$ & mentions of the name ``Robert.''\\
 & f9995 & $4.76$ $[4.14, 5.17]$ & references to specific individuals or names\\
 & f11250 & $3.62$ $[3.45, 4.54]$ & references to specific individuals or names\\
\midrule
\multicolumn{4}{l}{\textbf{f6240} (anchor, $H(a)=3.01$ bits): \textit{temporal references indicating events relative to each other}}\\
decoder & f6389 & $3.54$ $[3.11, 3.88]$ & references to distance and location\\
 & f4270 & $2.71$ $[2.22, 2.99]$ & terms related to time duration or intervals\\
 & f15504 & $3.42$ $[3.09, 3.58]$ & verbs that indicate development or progress in various contexts\\
\cmidrule(lr){2-4}
semantic & f14539 & $3.87$ $[3.62, 4.26]$ & temporal references concerning events and their sequences\\
 & f4358 & $4.33$ $[3.94, 4.69]$ & temporal references and dates related to events\\
 & f380 & $3.86$ $[3.38, 4.20]$ & temporal references to durations and past events\\
\midrule
\multicolumn{4}{l}{\textbf{f10360} (anchor, $H(a)=3.11$ bits): \textit{clauses that introduce relative clauses or provide additional information}}\\
decoder & f9363 & $3.46$ $[3.15, 3.95]$ & words that refer to or describe individuals, particularly in the context of actions and relationships\\
 & f13965 & $2.15$ $[2.00, 2.43]$ & possessive forms and statements of ownership or existence\\
 & f14078 & $4.22$ $[3.49, 4.60]$ & references to specific claims or statements about the nature of things\\
\cmidrule(lr){2-4}
semantic & f11355 & $4.59$ $[4.21, 4.87]$ & clauses that introduce additional information or elaborate on preceding statements\\
 & f7409 & $4.28$ $[3.90, 4.66]$ & conjunctions and relative clauses that indicate relationships or conditions\\
 & f13300 & $4.29$ $[3.99, 4.67]$ & conjunctions and phrases indicating additional information or contrasts\\
\midrule
\multicolumn{4}{l}{\textbf{f13603} (anchor, $H(a)=5.09$ bits): \textit{conditional language and uncertainty expressions}}\\
decoder & f796 & $5.30$ $[4.96, 5.52]$ & phrases related to medical procedures and their effectiveness\\
 & f11741 & $5.41$ $[4.94, 5.59]$ & inclusive language and sentiments inviting participation or collective experience\\
 & f1611 & $3.58$ $[3.20, 4.17]$ & emotional expressions and reactions related to disappointment and frustration\\
\cmidrule(lr){2-4}
semantic & f12287 & $3.20$ $[2.67, 3.50]$ & conditional statements and expressions of uncertainty\\
 & f6409 & $3.36$ $[2.83, 3.79]$ & conditional statements and expressions of uncertainty or potential outcomes\\
 & f3343 & $1.44$ $[1.27, 1.70]$ & conditional language indicating possibility or potential outcomes\\
\midrule
\multicolumn{4}{l}{\textbf{f6900} (anchor, $H(a)=5.10$ bits): \textit{mentions of competition and achievements}}\\
decoder & f11813 & $5.54$ $[5.23, 5.73]$ & structured data representations and definitions\\
 & f8958 & $5.13$ $[4.92, 5.37]$ & occurrences of vehicle-related crashes and their contexts\\
 & f9171 & $5.02$ $[4.57, 5.26]$ & references to serum levels and indicators of medical conditions\\
\cmidrule(lr){2-4}
semantic & f9816 & $3.83$ $[3.50, 4.52]$ & references to achievements and accolades in contexts related to performance or competitions\\
 & f7756 & $4.50$ $[4.19, 4.88]$ & references to competitions, awards, and sporting events\\
 & f119 & $3.88$ $[3.30, 4.44]$ & references to athletic achievements and performances\\
\bottomrule
\end{tabular}
\caption{For each anchor feature of Fig.~\ref{fig:fnl-clusters-semantic}, its three nearest neighbours by decoder cosine and its three nearest by auto-interp embedding cosine, with each feature's median nonlocality over its firing events and the interquartile range in brackets. The semantic neighbours are near-paraphrases of the anchor's description yet spread widely in nonlocality; the decoder neighbours need not share the anchor's topic yet sit close to it on the axis.}
\label{tab:supp-clusters}
\end{table*}

\subsection{Relation of FNL to token-cue and corpus-contrast selectors}
\label{sec:supp-selectors}

A natural objection to FNL is that it might be a repackaging of a selection signal already in use. We test it against the two candidates directly, on the same model and layer. Both candidates are activation-based feature-filtering methods from the reasoning-features literature discussed in the related work of the main text, and we include the definitions for completeness. ReasonScore \cite{galichin2025reasoning} scores each feature by its average activation mass on a curated vocabulary of reasoning cue words, so features firing on reasoning-flavoured tokens rank highly. 

Corpus-contrast enrichment compares a feature's firing rates on a reasoning corpus $R$ (OpenThoughts-114k) and a general chat corpus $C$ (LMSYS-Chat), both under the same chat template. Writing $r_a(\cdot)$ for the fraction of corpus tokens at which feature $a$'s activation exceeds a threshold $\tau=0.2$, the contrastive enrichment is $\mathrm{enrich}_a = r_a(R) / \max\!\big(r_a(C),\, 1/n_C\big)$, where $n_C$ is the number of tokens in $C$ and the regularization by $1/n_C$ in the denominator gives features that never fire on chat a finite score. A feature is reasoning-enriched when $\mathrm{enrich}_a \gg 1$, meaning it fires almost exclusively on reasoning text.

\begin{figure}[t]
    \centering
    \includegraphics[width=\columnwidth]{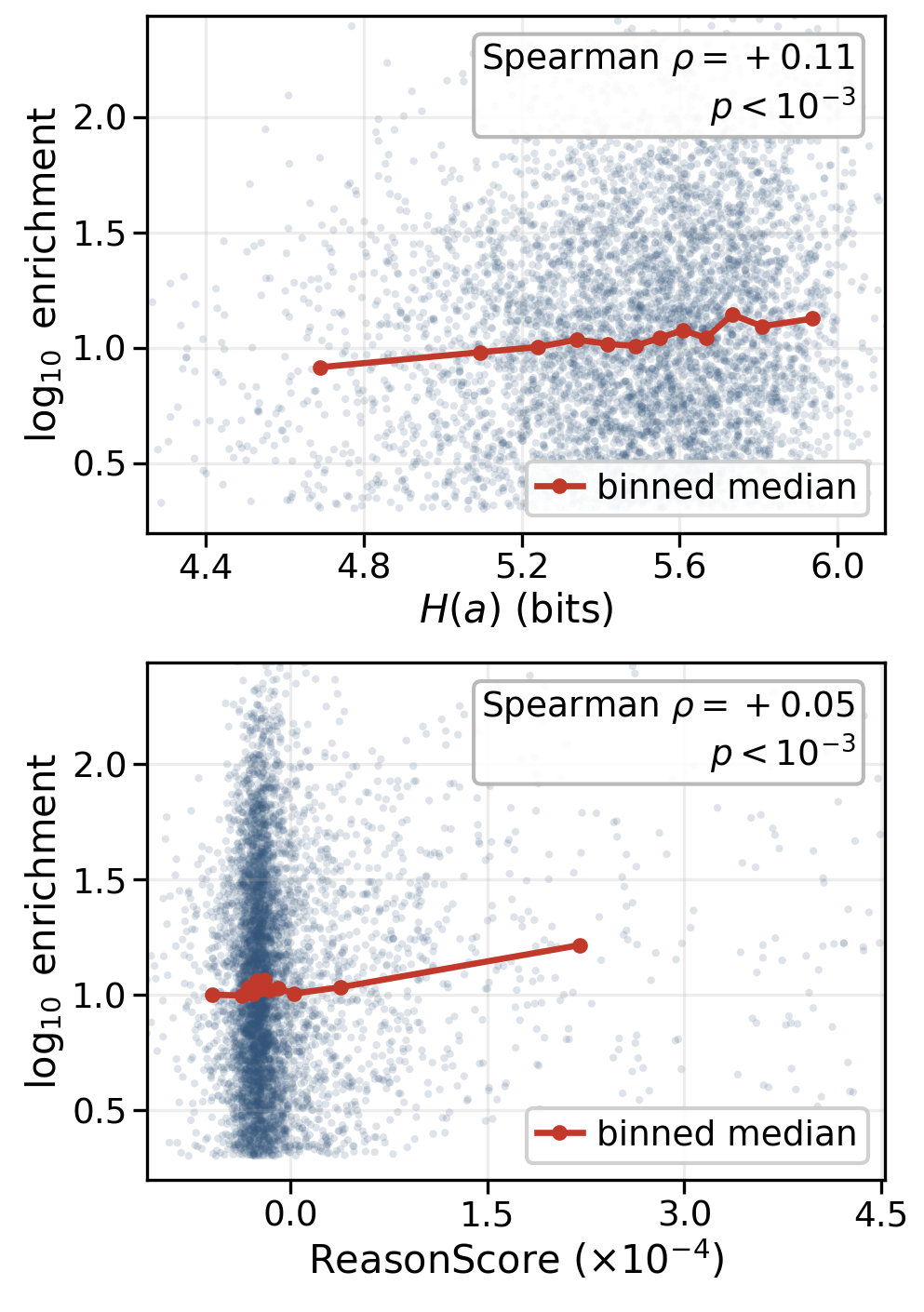}
    \caption{Contrastive enrichment against the two other selection axes, over the $5{,}358$ dictionary features carrying both a valid nonlocality and an enrichment score (DeepSeek-R1-Distill-Llama-8B layer 19; red curves are binned medians; axes clipped to the central $98\%$ of each variable). \emph{Top:} $\log_{10}$ enrichment against FNL. \emph{Bottom:} against ReasonScore. Nonlocality is a stronger correlator of the two with higher correlation and wider ranges, whereas for ReasonScore most features cluster at zero value.}
    \label{fig:fnl-selectors}
\end{figure}

For ReasonScore \cite{galichin2025reasoning} we compute the correlation between FNL and ReasonScore across $98$ features carrying both quantities, and found these two metric are close to being orthogonal with Spearman $\rho = -0.214$ ($p = 0.035$). This suggests FNL is not a token-frequency statistic in disguise, and a feature selected as reasoning-relevant by vocabulary is not thereby high in nonlocality. Corpus-contrast enrichment, on the other hand, shows a mild positive correlation with the feature nonlocality: the $212$ features enriched by a factor of $100$ or more average $5.56$ bits, against a dictionary-wide mean of $5.05$ bits. Figure~\ref{fig:fnl-selectors} plots enrichment against the two other axes over the full scored dictionary to visualize the correlations between those metrics.

\subsection{FNL and auto-interpretation}
\label{sec:supp-continuum}

\begin{figure*}[p]
    \centering
    \includegraphics[height=0.83\textheight]{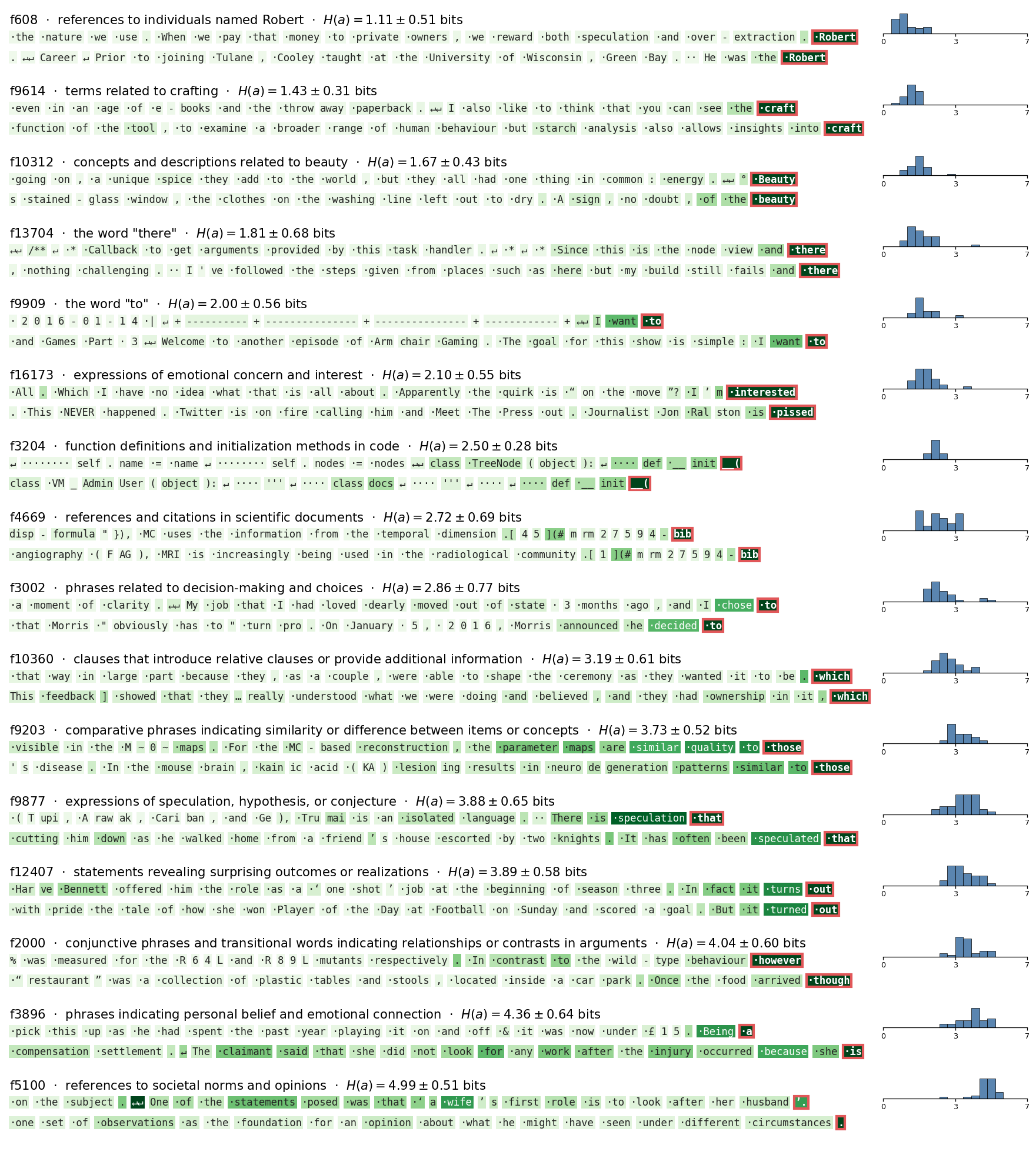}
    \caption{Sixteen Gemma-2-2B layer-12 features ordered by feature nonlocality, from $1.11$ to $4.99$ bits. Each block gives the feature index, its auto-interpretation label, and its mean nonlocality with the standard deviation across the feature's activating events, followed by two maximally activating passages. Within a passage, each token's background is shaded by its per-position influence $J_a(t)$ on the feature's activation (darker is more influential) and the activating token is boxed in red. Passages are truncated to the tail of the $128$-token measurement window for legibility. At the right of each block, a histogram gives the full distribution of that feature's nonlocality over its activating events, drawn on a common axis from $0$ to the ceiling $\log_2 128 = 7$ bits, so the distributions are directly comparable down the page. Influence is concentrated on the activating token itself at the top of the figure and spreads across the passage towards the bottom, and the histogram mass moves rightward correspondingly. Features f608 and f10360 also appear in Fig.~\ref{fig:feature-continuum} of the main text with identical values.}
    \label{fig:fnl-continuum-supp}
\end{figure*}

Figure~\ref{fig:feature-continuum} of the main text shows seven features and three influence maps to illustrate the association between verbal abstractness and feature nonlocality. Figure~\ref{fig:fnl-continuum-supp} expands this comparison to sixteen Gemma-2-2B layer-12 features ordered by nonlocality, each with its auto-interpretation label, the full distribution of its nonlocality across its firing events, and two of its maximally activating passages with every token shaded by its per-position influence and the activating token boxed.

\section{Envelope Steering Across Layers, Models, and Dictionaries}
\label{sec:supp-steering}

This section reports the full envelope-steering grid, extending the proof-of-concept experiments of the main text. The grid covers the following setups, with $\gamma$ the multiplicative gain applied to a selected feature's activation whenever it is active and ``high'' / ``low'' the envelope built from the highest- or lowest-FNL fraction of the dictionary.
\begin{itemize}
\item \emph{Models, dictionaries, layers:} DeepSeek-R1-Distill-Llama-8B with the 65k dictionary of \citet{galichin2025reasoning} at layer 19 and with Llama~Scope \cite{he2024llamascope} R1-distill dictionaries at layers 10--12; Gemma-2-9B (base) with Gemma~Scope \cite{lieberum2024gemmascope} at layers 9, 20, 31; Qwen3-8B (thinking) with Qwen-Scope \cite{deng2026qwenscope} at layers 9, 18, 27, after verifying that the base-trained dictionary transfers to the post-trained checkpoint.
\item \emph{Envelopes:} top, bottom, or size-matched random fraction ($20\%$, $10\%$, $5\%$) of the dictionary by FNL.
\item \emph{Steering gain:} $\gamma \in [1.05, 5]$, calibrated per layer on held-out tuning problems as the largest gain introducing no generation failures absent from the unsteered model via a pilot steering strength sweep on 20 held-out MATH-500 questions. All steering experiments use the clamping strategy explained in the main text.
\item \emph{Evaluation:} MATH-500 \cite{lightman2023letsverify,hendrycks2021math} with four sampled rollouts at $T{=}0.6$, majority vote of four, graded by a local Llama-3.3-70B judge; MMLU-Pro \cite{wang2024mmlupro} zero-shot chain of thought (majority of four) for the reasoning models and the published five-shot greedy protocol ($n{=}1400$) for Gemma, whose unsteered accuracy reproduces the published reference; all generation through the LM Evaluation Harness on vLLM.
\end{itemize}
Table~\ref{tab:supp-grid} collects the grid cells behind the two claims of the main text, and Table~\ref{tab:supp-ledger} the interval-backed contrasts.

We now report extensive steering experiment over the above setups as further validations for results in the main text. Firstly, feature nonlocality carries some selective power over an arbitrary envelope of the same size on the Llama-family reasoning model in improving reasoning performance, although the selection effect is weaker when transferring to other models. Table~\ref{tab:supp-ledger} shows that at layer 19 the high-FNL envelope beats the size-matched random envelope; the same contrast is positive but less robust at Llama~Scope layer 12 and runs negative on Gemma. Secondly, we find that steering high-FNL features is better than steering low-FNL features across the models we test. The high-vs-low contrast is significantly visible on the reasoning-distilled model and on Gemma, where low-FNL arm may collapse and degrade reasoning to random guessing at high steering strengths, while the high-FNL arm stays near baseline consistently. Similar results in Qwen3-8B are also reported in Table~\ref{tab:supp-grid}. The asymmetry is attributable to nonlocality rather than to activation statistics, since a control envelope selected by activation magnitude alone, injecting more perturbation than the low arm, is significantly less harmful.

\begin{table*}[t]
    \centering
    \small
    \setlength{\tabcolsep}{5pt}
    \begin{tabular}{lcccccccc}
        \toprule
        cell & fraction & $\gamma$ & $n$ & unsteered & high & low & random & high$-$low \\
        \midrule
        \multicolumn{9}{l}{\emph{DeepSeek-R1-Distill-Llama-8B, layer 19, 65k SAE; MATH-500, majority of 4}}\\
        L19 & 20\% & 1.10 & 480 & 0.881 & \textbf{0.942} & 0.923 & 0.940 & $+0.019$ \\
        L19 & 20\% & 1.10 & 100 & 0.850 & \textbf{0.930} & 0.870 & 0.910 & $+0.060$ \\
        L19 & 10\% & 1.20 & 100 & 0.850 & 0.910 & 0.890 & 0.933 & $+0.020$ \\
        L19 & \phantom{0}5\% & 1.40 & 100 & 0.850 & 0.930 & 0.920 & 0.923 & $+0.010$ \\
        \midrule
        \multicolumn{9}{l}{\emph{Same model, Llama Scope R1-distill dictionaries; MATH-500, majority of 4, shared unsteered control}}\\
        L10 & 10\% & 1.4 & 480 & 0.881 & 0.938 & 0.927 & 0.932 & $+0.011$ \\
        L11 & 10\% & 1.8 & 480 & 0.881 & 0.929 & 0.938 & 0.932 & $-0.009$ \\
        L12 & 20\% & 1.1 & 480 & 0.881 & \textbf{0.944} & 0.931 & 0.935 & $+0.013$ \\
        \midrule
        \multicolumn{9}{l}{\emph{Same model, layer 19; MMLU-Pro, majority of 4}}\\
        L19 & 20\% & 1.10 & 120 & 0.533 & 0.567 & 0.525 & 0.608 & $+0.042$ \\
        \midrule
        \multicolumn{9}{l}{\emph{Gemma-2-9B (base), Gemma Scope; MMLU-Pro, greedy}}\\
        L9  & 20\% & 2 & 1400 & 0.440 & 0.436 & 0.432 & 0.431 & $+0.004$ \\
        L20 & 20\% & 2 & 1400 & 0.440 & 0.424 & 0.388 & 0.437 & $+0.036$ \\
        L20 & 10\% & 3 & 1400 & 0.440 & 0.413 & 0.321 & 0.424 & $+0.092$ \\
        L20 & \phantom{0}5\% & 5 & 1400 & 0.440 & 0.407 & 0.083 & 0.415 & $+0.324$ \\
        \midrule
        \multicolumn{9}{l}{\emph{Qwen3-8B (thinking), Qwen-Scope layers 9/18/27 of 36; MMLU-Pro, majority of 4, shared unsteered control}}\\
        L9  & 20\% & 1.2 & 600 & 0.752 & 0.752 & 0.740 & 0.743 & $+0.012$ \\
        L18 & 20\% & 1.1 & 600 & 0.752 & 0.747 & 0.752 & 0.755$^{\dagger}$ & $-0.005$ \\
        L27 & 20\% & 2.0 & 600 & 0.752 & 0.753 & 0.745 & 0.737$^{\dagger}$ & $+0.008$ \\
        \bottomrule
    \end{tabular}
    \caption{Envelope-steering cells supporting the two claims of the main text, selected from the full sweep. ``High / ``low are the envelopes over the highest- / lowest-FNL fraction of the dictionary; ``random is the mean over 3--10 size-matched random draws, except where marked $^{\dagger}$ (a single draw, on dose-curve rows that reuse the primary cell's control). $n$ is the number of scored problems per condition; bold marks cells where the high-FNL envelope is the best-scoring condition. The panels divide by whether the unsteered model degenerates: on the two R1-distilled stacks every high-FNL cell beats the unsteered control by $+4.6$ to $+8.0$ points, while on the base model and the native reasoning model (lower panels), where the unsteered model has no degenerate failure mode, no condition beats the control. The low-FNL envelope on Gemma Scope is the only arm that is significantly harmful.}
    \label{tab:supp-grid}
\end{table*}

\begin{table*}[t]
    \centering
    \small
    \setlength{\tabcolsep}{3.5pt}
    \begin{tabular}{llcc}
        \toprule
        contrast & cell & $\Delta$ & 90\% CI \\
        \midrule
        \multicolumn{4}{l}{\emph{high $-$ random (claim 1: selective power over an arbitrary envelope)}}\\
        avg@4 & L19, 20\%, $n{=}480$ & $+0.010$ & $[\phantom{+}0.000, +0.021]$ \\
        maj@4 & L12, 20\%, $n{=}480$ & $+0.009$ & $[-0.001, +0.019]$ \\
        greedy & Gemma L20, 20\% & $-0.013$ & $[-0.024, -0.002]$ \\
        \midrule
        \multicolumn{4}{l}{\emph{high $-$ low (claim 2: high-FNL steering beats low-FNL steering)}}\\
        maj@4 & L19, 20\%, $n{=}100$ & $+0.060$ & $[+0.020, +0.100]$ \\
        maj@4 & L19, 20\%, $n{=}480$ & $+0.019$ & $[+0.004, +0.035]$ \\
        greedy & Gemma L20, 20\% & $+0.036$ & $[+0.018, +0.055]$ \\
        \midrule
        \multicolumn{4}{l}{\emph{low-FNL harm beyond activation magnitude (control for claim 2)}}\\
        greedy & Gemma L20, 20\% & $-0.023$ & $[-0.039, -0.008]$ \\
        \bottomrule
    \end{tabular}
    \caption{Statistical significance illustration for the two claims in the main text. maj@4 / avg@4 are majority-vote and mean accuracy over four rollouts; ``greedy'' is single-sample greedy accuracy ($n{=}1400$). Every interval is a $10{,}000$-resample paired bootstrap over problems at a fixed seed; contrasts against the random envelope additionally resample the random draws.}
    \label{tab:supp-ledger}
\end{table*}

\section{Reproduction Details of Reference Papers}
\label{sec:supp-repro-refs}

\subsection{Token-injection pipeline of \citet{ma2026injection}}
\label{sec:supp-injection-repro}

Because \citet{ma2026injection} release code but no per-feature results, we re-ran their pipeline in full (contrastive selection of the top 100 features, eight injection strategies per feature, classification by the best strategy's effect size) and reproduce their published classification nearly exactly: class counts of $46/12/21/21$ against their $46/12/19/23$, with matching injection effect sizes and winning-strategy fractions. The only quantity that does not reconcile is the detection-stage effect size, which we attribute to a normalization difference in their reporting, since every quantity downstream of it agrees.

\subsection{Steering defence of \citet{assogba2026sae}}
\label{sec:supp-ccdelta-repro}

Before auditing the jailbreaking-mitigation features, we reproduced their CC-delta pipeline and verified their out-of-distribution jailbreaking mitigation result. Against the held-out Few-Shot-JSON attack the sparse feature-steering direction raises safety from $0.511$ unsteered to $0.753$ at $\lVert v \rVert = 1.6$ and $0.895$ at $\lVert v \rVert = 3.2$, over the same $404$ requests, while a dense contrastive-activation direction at matched strength reaches only $0.343$ and a random direction $0.567$. 

\section{Reproducibility Details}
\label{sec:repro}

\subsection{Computing infrastructure}

All experiments ran on single compute nodes of an HPC cluster. GPU jobs used NVIDIA H200 NVL (141\,GB), A100 80\,GB PCIe, RTX 6000 Ada (48\,GB), and RTX 4090 (24\,GB); the sub-2B models fit on the 24\,GB card, the 8B--9B models with $10^4$--$10^5$-feature dictionaries ran on the H200 or A100, and benchmark generation used three or four GPUs per job, one or two of them serving the grading model. The recorded total runtime is on the order of $1{,}000$ GPU-hours, dominated by the dictionary-scale nonlocality scans and the steered benchmark generation.

\subsection{Datasets}
No novel dataset is introduced. The text corpora used include WikiText \cite{merity2016pointer}, GSM8K \cite{cobbe2021gsm8k}, Python code from The Stack \cite{kocetkov2022stack}, FineWeb-Edu \cite{lozhkov2024fineweb}, The Pile \cite{gao2020pile}, OpenThoughts-114k \cite{openthoughts2025}, and LMSYS-Chat-1M \cite{zheng2023lmsyschat}. For benchmarks we used MATH-500 \cite{lightman2023letsverify,hendrycks2021math}, MMLU-Pro \cite{wang2024mmlupro}, and the $404$-request jailbreak evaluation set with its held-out Few-Shot-JSON wrapper \citep{assogba2026sae}.

\subsection{Runtimes and number of runs}

Wall-clock times below are the recorded maxima for each job family, taken from the run logs. They are single-node figures on the hardware described above and are intended to let a reader budget a replication.

\emph{Measurement.} A dictionary-scale nonlocality scan is the dominant cost. Over a full $10^{4}$--$10^{5}$-feature dictionary would be around 7-8 hours on four H200 cards. The depth sweep costs $4.5$--$5.5$ hours per layer for Gemma-2-2B.

\emph{Generation and grading.} Steered benchmark generation for MATH-500 takes around 4 hours for 7B-8B models

\emph{Number of runs.} Nonlocality values are means over $16$--$60$ firing events per feature, stated per experiment and default to $k=32$. Benchmark accuracies on the reasoning models are means or majority votes over four sampled rollouts per problem; Gemma cells are single-sample greedy. Random-envelope arms are $3$--$10$ independent draws, pooled per problem, with the count given per cell in Table~\ref{tab:supp-grid}.

\subsection{Random seeds}

Four distinct sources of randomness enter the results, and each is seeded from an explicit command-line argument with a fixed default, recorded in the output file of every run.

\emph{Corpus sampling.} Which documents and which firing events enter a nonlocality measurement is drawn from a NumPy generator seeded per experiment (defaults 260705, 260706, 260720 for the injection, paraphrase and cross-corpus runs respectively).

\emph{Envelope construction.} The size-matched random envelopes are drawn from a generator seeded at 260622; where a cell reports a mean over several random draws, draw $k$ uses seed $260622+k$, so the reported spread is over reproducible draws rather than over an unrecorded sequence.

\emph{Benchmark decoding.} Every generation arm passes a fixed engine seed (260621) to vLLM, so the four sampled rollouts per problem are reproducible given the same engine version and GPU count. We note the standard caveat that batched GPU inference is not bit-reproducible across different hardware or batch schedules even at a fixed seed, so exact per-rollout texts may differ on a different allocation while the aggregate accuracies do not.

\emph{Bootstrap resampling.} All confidence intervals are paired bootstraps over problems with $10{,}000$ resamples at seed 260622 (260621 for the majority-vote analyses), so the reported interval endpoints are exactly reproducible from the stored per-problem scores without rerunning any generation.

\subsection{Hyperparameters: values tried and how the final ones were chosen}
Table~\ref{tab:repro-hparams} lists the hyperparameter choices. We note that this table is for bookkeeping as the specifications have been stated near each experiment. We record the range explored, the value used for the reported results if selected, and the criterion that guides the selection if applicable, across the experiments conducted. The steering gain is the only hyperparameter selected against a performance signal, calibrated on held-out tuning problems disjoint from the reported ones.

\begin{table}[t]
\centering
\small
\setlength{\tabcolsep}{2pt}
\begin{tabular}{lll>{\raggedright\arraybackslash}p{0.26\columnwidth}}
\toprule
Quantity & Range tried & Used & Criterion \\
\midrule
Context window $T$      & 8--2048 & 128 / 64 &  approximate saturation of $H$ \\
Events/feature $k$      & 8--60   & 32 / 60  & estimator stability \\
Firing thresh.\ $\tau$  & 0, 0.2  & 0, 0.2    & SAE gate, ReLU or JumpReLU \\
\midrule
Envelope fraction       & 5/10/20\% & 5/10/20\% & N/A \\
Steering gain $\gamma$  & 1.05--5 & per cell & largest gain with no new failures on the held-out pilot \\
\midrule
Decoding temp.\          & 0/0.6/0.8 & 0.6 & model card \\
Decoding top-$p$        & 0.95, 1 & 0.95 & model card \\
Rollouts per problem    & 1, 4    & 4    & majority vote \\
Max gen.\ tokens        & 16{,}384 & 16{,}384 & no truncation \\
\bottomrule
\end{tabular}
\caption{Hyperparameters varied during development, with the value used for the reported results and the criterion that fixed it. The first group record the general settings used when computing feature nonlocalities. The second group include the steering experiments setups, and the last group contains details on rollout generations and evaluation procedures.}
\label{tab:repro-hparams}
\end{table}

\subsection{Beginning-of-sequence convention}
\label{sec:supp-protocol}

The nonlocality measurements in this work cut windows without a beginning-of-sequence token and exclude the window-initial position, which hosts the attention sink, from the influence support; both choices shift absolute values without disturbing any ordering the paper uses, so absolute values are comparable only across studies adopting the same convention.

\bibliography{references}
\end{document}